\documentclass[letterpaper, 10 pt, conference]{ieeeconf}  % Comment this line out if you need a4paper

\IEEEoverridecommandlockouts                              % This command is only needed if 
\usepackage{graphics} % for pdf, bitmapped graphics files
\usepackage{epsfig} % for postscript graphics files
\usepackage{times} % assumes new font selection scheme installed
\usepackage{amsmath} % assumes amsmath package installed
\usepackage{amssymb}  % assumes amsmath package installed
\usepackage[backend=biber,sorting=nyt,sortcites=true]{biblatex}
\usepackage{multirow}
\usepackage{multicol}
\usepackage{hyperref}
\usepackage{booktabs}
\usepackage{enumerate}
\usepackage{color}
\usepackage{tabularx}
\usepackage{hyperref}
\usepackage{pifont}
\usepackage{algorithm}
\usepackage{algpseudocode}
\usepackage{amsmath}
\usepackage{adjustbox}
\usepackage[usenames,dvipsnames,table]{xcolor}
\hypersetup{
    colorlinks=true,
    linkcolor=MidnightBlue,
    filecolor=magenta,      
    urlcolor=MidnightBlue,
    citecolor=MidnightBlue,
} 
\usepackage[font=small,labelfont=bf]{caption}
\usepackage{marvosym}

\definecolor{darkred}{rgb}{0.76, 0.23, 0.13}
\definecolor{darkgreen}{rgb}{0.01, 0.75, 0.24}
\definecolor{darkgray}{rgb}{0.66, 0.66, 0.66}

\title{\LARGE \bf
Rethinking Visual Embodiment Dependence in Visuomotor Policies
}

\author{
Hongjie Fang$^{1,2,3}$, Yuxuan Lu$^{1}$, Chenxi Wang$^3$, Haoxiang Qin$^{1}$, Shirun Tang$^{2,3}$, \\ Zihao He$^1$, Shangning Xia$^3$, Jingjing Chen$^1$, Wanxi Liu$^{2,3,4}$, Shiquan Wang$^{2,3,4,\dagger}$, Cewu Lu$^{1,3,4,5,\dagger}$%
\thanks{$^\dagger$Corresponding Authors. $^1$Shanghai Jiao Tong University. $^2$FORTE Lab. $^3$Noematrix. $^4$Flexiv. $^5$Shanghai Innovation Institute.}
}

\begin{document}

\makeatletter
\let\@oldmaketitle\@maketitle% Store \@maketitle
\renewcommand{\@maketitle}{
\@oldmaketitle% Update \@maketitle to insert...
\vspace{0.2cm}
\centering
\includegraphics[width=\linewidth]{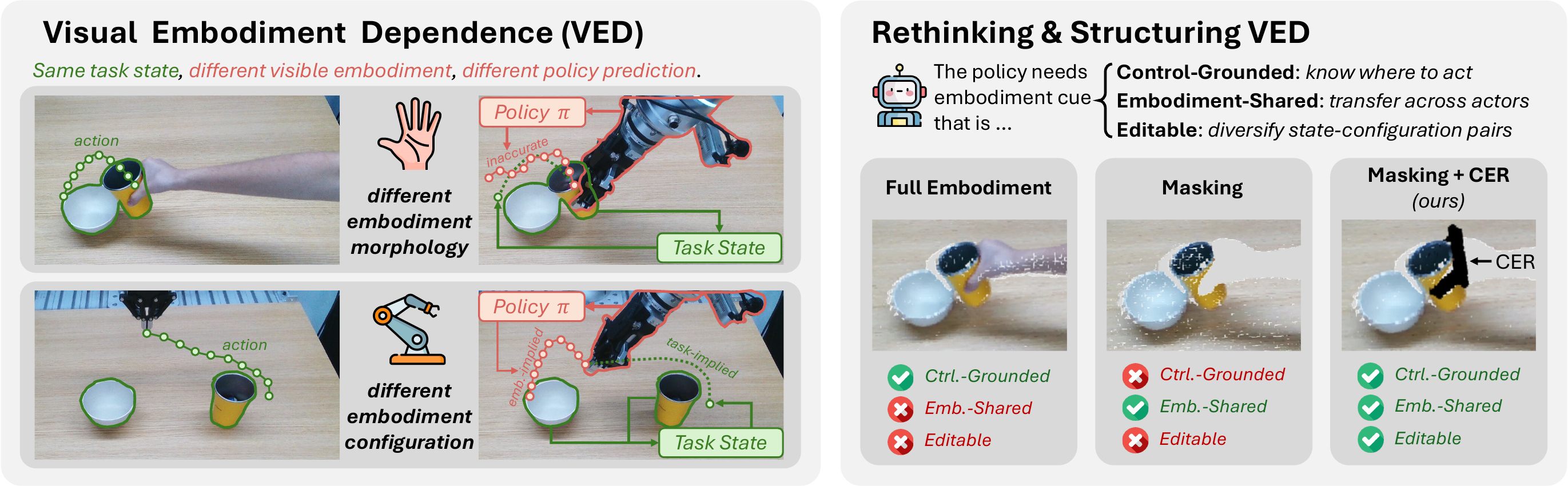}

\captionof{figure}{
\textbf{Rethinking Visual Embodiment Dependence (VED).}
\textit{\textbf{(Left)}} The same task state can induce different policy predictions when visible embodiment morphology or configuration changes.
\textit{\textbf{(Right)}} We structure VED through 3D embodiment canonicalization with a CER that is control-grounded, shared across embodiments, and editable, enabling human-to-robot policy transfer without robot demonstrations and robust generalization to unfamiliar robot configurations.
}
\label{fig:teaser}\vspace{-0.45cm}
}%
\makeatother

\maketitle
\thispagestyle{empty}
\pagestyle{empty}
\addtocounter{figure}{-1}

%%%%%%%%%%%%%%%%%%%%%%%%%%%%%%%%%%%%%%%%%%%%%%%%%%%%%%%%%%%%%%%%%%%%%%%%%%%%%%%%
\begin{abstract}
Visuomotor policies observe both the task scene and the acting embodiment, allowing embodiment-specific visual cues to influence action prediction. We study this phenomenon as visual embodiment dependence (VED) and show, through cue-conflict interventions across representative policies, that visible robot configuration can become a shortcut to task progress. Rather than eliminating VED, we argue that it should be structured around embodiment information that supports control and generalization. We realize this through embodiment canonicalization in 3D point clouds, replacing the original embodiment with a canonical end-effector representation (CER) that preserves control-relevant geometry while abstracting embodiment-specific morphology. Its editable form further enables configuration-decorrelation augmentation for unfamiliar robot configurations. Experiments show that embodiment canonicalization substantially improves human-to-robot policy transfer without robot demonstrations, while simply removing the embodiment is insufficient without preserving control-relevant geometry. We further find that CER itself can become a configuration shortcut when robot configuration becomes decoupled from task progress; configuration-decorrelation augmentation mitigates this failure mode and restores robust recovery without sacrificing performance on seen configurations. Together, these results show that robust visuomotor learning benefits from structuring, rather than removing, visual embodiment information. Project website: \href{https://tonyfang.net/ved}{https://tonyfang.net/ved}.
\end{abstract}

\section{Introduction}

Visuomotor policies have shown strong potential for learning manipulation directly from visual demonstrations~\cite{act, dp, dp3, rise, airexo2_rise2, pi0, pi05, oxe, gr00t, rt1, rt2, act3d, openvla, octo}. However, their visual observations typically contain not only the task scene, but also the acting embodiment itself, making embodiment appearance, geometry, and configuration readily available for action prediction. Cross-embodiment learning makes the consequences of such reliance particularly visible, motivating aligned data collection devices~\cite{umi,stick,legato,fastumi,demoat,dobbe_stickv1} and observation canonicalization methods~\cite{airexo2_rise2,masquerade,phantom,dexumi,shadow,mirage}. While they primarily address discrepancies between different embodiments, similar dependence may also matter within a fixed embodiment when its visible configuration becomes decoupled from the task state in ways not seen during training. This raises a broader question: \textit{how do visuomotor policies rely on the visible embodiment, and how does this dependence affect generalization?}

We refer to the dependence of policy predictions on the visually observed embodiment as \textbf{visual embodiment dependence (VED)}. VED is not inherently undesirable: visual information about the end effector provides important spatial and interaction grounding for manipulation. The challenge arises when \textit{policies rely on embodiment cues that do not generalize}, which we consider in two forms. (1)~Morphology dependence: policies associate embodiment-specific morphological features with demonstrated actions, limiting transfer when the embodiment changes~\cite{phantom,masquerade}. (2)~Configuration dependence: policies use visible embodiment configuration as a proxy for task progress when the two are strongly correlated in successful demonstrations, causing errors once they become decoupled~\cite{xie2026s,lu2026would, propriovla}. Thus, our goal is \textit{not to eliminate VED}, but to \textit{structure it around embodiment information that supports control and generalization}.

We next directly examine the second form of VED, configuration dependence, which remains less explored within the same embodiment. We construct cue-conflict observations that pair the task state from one stage with the visible embodiment configuration from another, and measure whether predicted actions follow the \textit{task-implied} or \textit{embodiment-implied} behavior. Across four representative policies in simulation and the real world, we find pronounced configuration dependence at specific task states: when visible robot configuration strongly correlates with demonstrated task progress, it can dominate task-state cues and drive predictions toward the \textit{embodiment-implied} action, as shown in Fig.~\ref{fig:teaser}(left).

These observations suggest that useful visual embodiment information should be \textit{control-grounded}, \textit{shared across embodiments}, and \textit{editable}, as illustrated in Fig.~\ref{fig:teaser}(right). We realize these properties through embodiment canonicalization in 3D point clouds: we mask the original embodiment geometry and replace it with a \textit{canonical end-effector representation (CER)}. In this work, our CER design takes the form of a canonical gripper that preserves control-relevant geometry while abstracting away embodiment-specific morphology. The same design can be instantiated from either human hand or robot gripper poses. Its editable 3D form further enables \textit{configuration-decorrelation augmentation}, which pairs off-trajectory CER configurations with corresponding recovery behaviors to diversify configuration-behavior associations.

Since our embodiment canonicalization operates in 3D, we use RISE~\cite{rise} as the visuomotor backbone. We evaluate our approach in two complementary settings. (1) In human-to-robot policy transfer, using only human demonstrations and no robot demonstrations for training, embodiment canonicalization consistently improves policy performance across diverse manipulation tasks. (2) When generalizing to novel robot configurations, configuration-decorrelation augmentation substantially improves off-trajectory recovery while preserving policy performance on seen configurations. Together, these results show that robust visuomotor learning benefits from structuring VED around shared, control-relevant geometry rather than embodiment-specific morphology or trajectory-correlated configuration cues.
\section{Related Work}\label{sec:related}

\subsection{Visuomotor Policy Learning}

Visuomotor policies learn manipulation behaviors by mapping visual observations to robot actions through imitation learning~\cite{bc}. Existing approaches broadly differ in how they represent visual observations. Image-based policies operate directly on RGB inputs and have been widely adopted for robotic manipulation~\cite{robomimic,act,dp,cage,spawnnet}. Recent vision-language-action (VLA) models~\cite{rt1,rt2,octo,openvla,pi0,pi05,oxe,gr00t,mtact} further leverage large-scale pre-training and multi-task robot datasets~\cite{oxe,rh20t,droid,robomind} to improve semantic and behavioral generalization. In parallel, 3D policies~\cite{peract,rvt,act3d,dp3,rise,airexo2_rise2,histrise,lift3d} encode point clouds, voxels, or 3D-aligned visual features, providing explicit geometric structure for spatial reasoning and viewpoint robustness.

Regardless of representation, visual observations typically contain both the task scene and the acting embodiment, making embodiment-specific morphology and configuration directly available for action prediction. Prior work~\cite{state,propriovla} has shown that proprioceptive robot states can become strongly correlated with demonstrated actions and form shortcuts that impair generalization. Our work studies the corresponding dependence on the \textit{visually observed embodiment}, which we term VED. Our approach operates in 3D point clouds, where embodiment geometry can be spatially isolated and manipulated independently of the surrounding task scene.

\subsection{Cross-Embodiment Policy Learning}
\label{sec:related-cross-emb}

Cross-embodiment policy learning must bridge differences between the embodiments observed in demonstrations and those used at deployment~\cite{cage}, a challenge that is particularly prominent in human-to-robot policy transfer~\cite{human_video_survey}.
Existing approaches reduce this gap in several ways.
(1) \textit{Interface-alignment} methods use handheld devices with wrist-centric sensing to make demonstration and deployment observations more consistent~\cite{umi,stick,legato,demoat,fastumi,dobbe_stickv1}.
(2) \textit{Observation-translation} methods transform human observations toward robot-like appearances~\cite{airexo2_rise2,masquerade,phantom,dexumi,mirage}.
(3) \textit{Shared-domain} approaches remove embodiment-specific regions to construct observations that are more consistent across embodiments~\cite{chang2023look,shadow,lidea}.
(4) \textit{Representation-abstraction} methods construct shared representations using keypoints or object-centric states~\cite{point_policy,kat,kalm,activeglasses,kdil,motion_track}.

Our work is most closely related to (3) and (4), but differs in the scope of abstraction: it preserves the full 3D task scene while applying \textit{embodiment canonicalization} only to the acting embodiment, replacing embodiment-specific geometry with our CER design while retaining control-relevant gripper geometry. Beyond cross-embodiment morphology, the visible CER configuration can still become strongly correlated with task progress. Its editable 3D form therefore enables configuration-decorrelation augmentation, which diversifies configuration-behavior associations during training.

\subsection{Generalization and Robustness in Visuomotor Policies}
\label{sec:related-generalizable}

Prior work has studied visuomotor generalization under variations in objects, backgrounds, camera viewpoints, and environments~\cite{cage,airexo2_rise2}. Large-scale robot datasets and policy pre-training broaden the behaviors and environments covered during training~\cite{mtact,rt1,oxe,openvla,octo,rt2,gr00t,pi05,hpt}; visual foundation models improve robustness to semantic and appearance variations~\cite{spawnnet,soft,cage,same}; and explicit 3D representations provide geometric priors for spatial and viewpoint generalization~\cite{act3d,rvt,rise,dp3,airexo2_rise2}. Data augmentation and synthetic demonstration generation further expand the observation and trajectory distributions available during training~\cite{mimicgen,cyberdemo,demogen}.

A complementary challenge arises when visible robot configuration becomes decoupled from task progress, as often occurs after execution errors or during recovery~\cite{armada,error1,error2,AgiaSinhaEtAl2024}. Our VED analysis studies this configuration dependence, while configuration-decorrelation augmentation uses editable CER to diversify configuration-behavior associations and improve policy robustness to such mismatches.
\section{Diagnosing Visual Embodiment Dependence}
\label{sec:ved_diagnostic}

\begin{figure}[t]
    \centering
    \includegraphics[width=\linewidth]{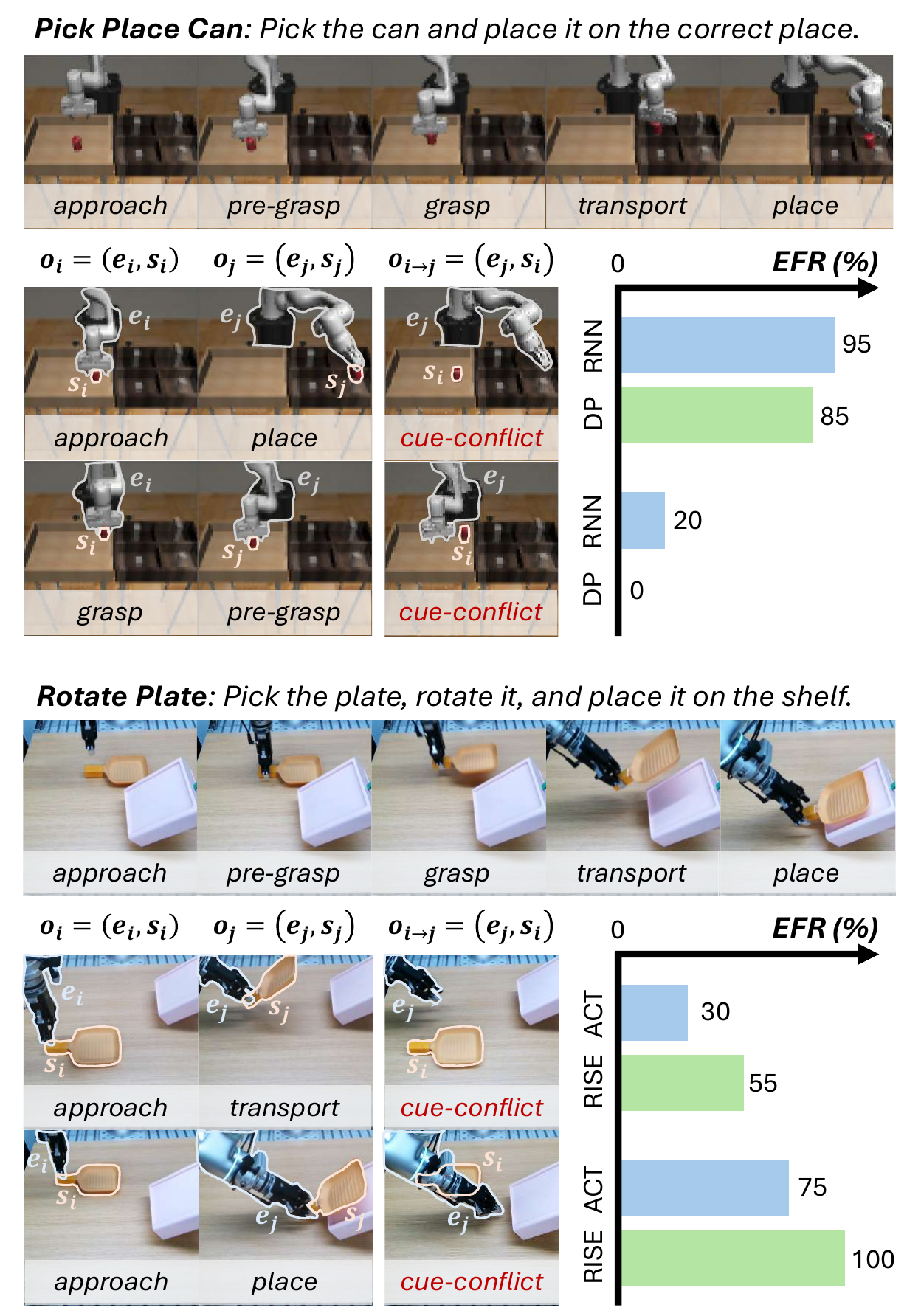}
\caption{
\textbf{Diagnosing Configuration Dependence in VED.}
We construct cue-conflict observations in simulated \textbf{\textit{Pick Place Can}} and real-world \textbf{\textit{Rotate Plate}} by pairing task state from one stage with embodiment configuration from another. Across four visuomotor policies, embodiment-following behavior varies substantially across task stages, revealing state-dependent but often pronounced reliance on visible robot configuration.
}
    \label{fig:diagnostic}\vspace{-0.4cm}
\end{figure}

Cross-embodiment literature~\cite{airexo2_rise2,cage,lidea,mirage,phantom} has extensively studied morphology differences across embodiments. We focus here on a less explored form of VED: \textit{configuration dependence within the same embodiment}. In successful demonstrations, robot configuration is often strongly correlated with task progress, making it a convenient proxy for the current task stage. We therefore construct controlled conflicts between configuration and task-state cues and examine which one governs the predicted action.

\subsection{Cue-Conflict Diagnostic}

Consider observations from two task stages: $o_i=(E_i,S_i)$ and $o_j=(E_j,S_j)$, where $E$ denotes the visible embodiment configuration and $S$ denotes the remaining visual task state. We construct a \textbf{cue-conflict observation} $o_{i\rightarrow j}=(E_j,S_i)$, which preserves task state $S_i$ while presenting the embodiment configuration from stage $j$. The task state therefore implies action $a_i$, whereas the visible configuration implies $a_j$ through the correlations present in the demonstrations.

In simulation, we restore the task state from stage $i$, set the robot to the configuration recorded at stage $j$, and re-render the observation. In the real world, we similarly reproduce the task state at stage $i$, move the robot to the configuration associated with stage $j$, and capture a new observation. This produces physically consistent observations while deliberately altering the correspondence between robot configuration and task progress. For each selected stage pair, we construct $N=20$ cue-conflict observations and evaluate policies on the same set. We include only stage pairs whose task- and embodiment-implied absolute actions are sufficiently distinct.

We introduce two controls to isolate visual configuration dependence. First, all diagnostic policies predict \textit{absolute actions}. Thus, changing the current robot configuration does not additionally transform a demonstrated target into an unseen relative displacement. Second, policies receive no proprioceptive robot state, such that configuration is available only through vision~\cite{rise, airexo2_rise2}. We additionally evaluate the corresponding observations $o_i$ and $o_j$ as a sanity check for general action-prediction errors, without filtering diagnostic samples based on performance.

Let $\hat a_{i\rightarrow j}=\pi(o_{i\rightarrow j})$, and let $s$ denote the current robot state expressed in the same task-space coordinates as the absolute actions. Since $a_i$ and $a_j$ may lie at different distances from $s$, directly comparing their distances to $\hat a_{i\rightarrow j}$ can be biased. We therefore measure the progress of the predicted action toward the two references:
\begin{equation}
\left\{
\begin{aligned}
G_E &= d(s,a_j)-d(\hat a_{i\rightarrow j},a_j),\\
G_T &= d(s,a_i)-d(\hat a_{i\rightarrow j},a_i),
\end{aligned}
\right.
\end{equation}
where $G_E$ and $G_T$ denote progress toward the \textit{embodiment-implied} and \textit{task-implied} actions, respectively. We define the relative embodiment progress as $\Delta G = G_E-G_T$. A positive $\Delta G$ indicates that the prediction makes greater progress toward the embodiment-implied action. We summarize this tendency using the \textbf{Embodiment-Following Rate (EFR)}:
\begin{equation}
\mathrm{EFR}=\frac{1}{N}\sum_{n=1}^{N}
\mathbf{1}\!\left[\Delta G^{(n)}>0\right].
\end{equation}
Higher EFR therefore indicates stronger reliance on visible embodiment configuration over task-state evidence under cue conflict. For absolute end-effector actions, $d(\cdot,\cdot)$ combines Euclidean translation error and geodesic rotation error, with rotation scaled to a comparable spatial scale.

\subsection{Configuration Dependence Analysis}

\begin{figure*}[t]
    \centering
    \includegraphics[width=\linewidth]{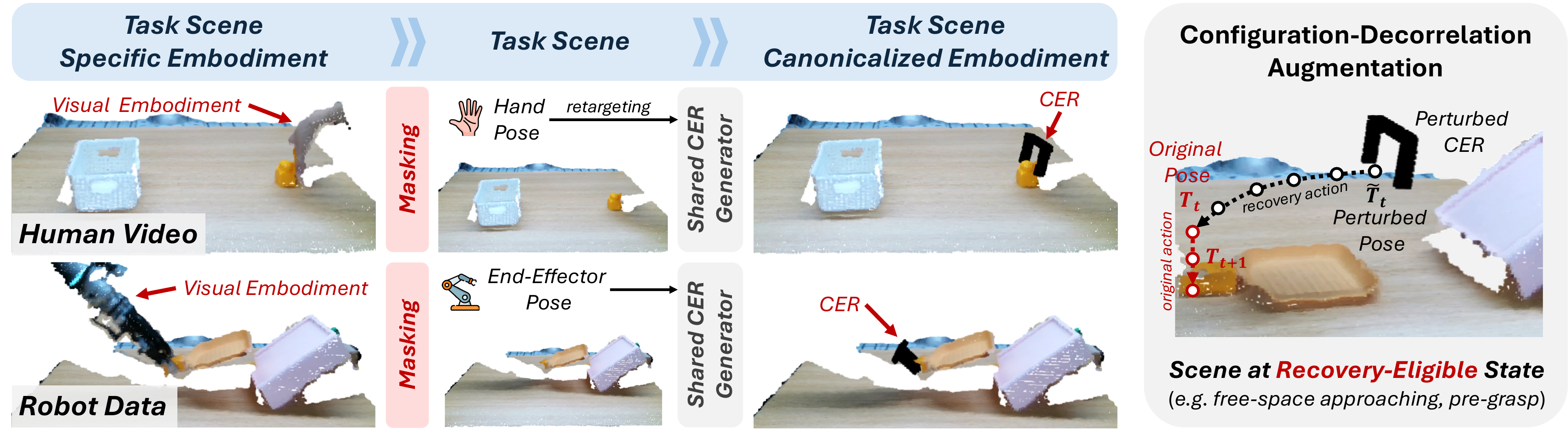}\vspace{-0.1cm}
\caption{
\textbf{Overview of Embodiment Canonicalization and Configuration-Decorrelation Augmentation.}
\textit{\textbf{(Left)}} Embodiment canonicalization removes the original embodiment geometry and replaces it with our CER design.
\textit{\textbf{(Right)}} CER can be directly edited in 3D to construct off-trajectory configurations paired with corresponding recovery behaviors for configuration-decorrelation augmentation.
}
    \label{fig:method}\vspace{-0.4cm}
\end{figure*}

We evaluate BC-RNN~\cite{robomimic} and DP~\cite{dp} in simulated \textbf{\textit{Pick Place Can}}, and ACT~\cite{act} and RISE~\cite{rise} in real-world \textbf{\textit{Rotate Plate}}. Fig.~\ref{fig:diagnostic} reports phase-wise EFRs under two representative cue-conflict scenarios for each task. The results reveal that configuration dependence is strongly \textit{state-dependent}. In \textbf{\textit{Pick Place Can}}, BC-RNN and DP strongly follow embodiment cues in the approach/transport conflict, whereas this tendency largely disappears in the grasp/pre-grasp conflict. A similar phase-dependent pattern appears in \textbf{\textit{Rotate Plate}} across ACT and RISE. These consistent patterns across architectures suggest that configuration dependence is not simply a consequence of weak policy performance, but becomes pronounced when visible robot configuration strongly predicts demonstrated task progress.

These results show that VED extends beyond morphology differences across embodiments: even within a fixed embodiment, visible configuration can act as a shortcut to task progress and bias predictions toward the embodiment-implied action. This motivates structuring VED to preserve control-relevant embodiment information while avoiding embodiment-specific morphology and configuration cues that are overly correlated with task progress. We realize this principle through \textit{embodiment canonicalization} with CER and \textit{configuration-decorrelation augmentation}.
\section{Method}
\label{sec:method}

\subsection{Overview}
\label{sec:method-overview}

We structure VED via two operations in 3D point clouds, as illustrated in Fig.~\ref{fig:method}. (1) \textit{Embodiment canonicalization} removes the original embodiment geometry and replaces it with CER, preserving control-relevant geometry while abstracting away embodiment-specific morphology (\S\ref{sec:method-cer}). (2) \textit{Configuration-decorrelation augmentation} exploits the editable CER to pair off-trajectory configurations with corresponding recovery behaviors, diversifying associations between task state and embodiment configuration during training (\S\ref{sec:method-aug}). Since both operations are defined directly in 3D, we instantiate our method with the 3D policy RISE~\cite{rise}.

\subsection{Embodiment Canonicalization}
\label{sec:method-cer}

Given a point cloud $P_t$ and embodiment mask $M_t$, let $P_t^{\mathrm{emb}}\subset P_t$ denote the points belonging to the embodiment. We remove these points to obtain the task-scene point cloud
\begin{equation}
    P_t^{\mathrm{scene}}
    =
    P_t \setminus P_t^{\mathrm{emb}}.
\end{equation}

We then replace the removed embodiment with CER. We define a canonical end-effector template with state $s$
\begin{equation}
    \mathcal{C}_0(s)=\{c_k\}_{k=1}^{K},
\end{equation}
represented by a set of 3D points in a canonical local frame. In this work, we focus on gripper-based manipulation and instantiate $\mathcal{C}_0(w_t)$ as a simple line-based canonical gripper with width $w_t$, as illustrated in Fig.~\ref{fig:method}. Specifically, two distal line segments represent the two gripper tips, whose separation encodes the gripper opening width, while a third proximal segment represents the gripper body and provides additional geometric orientation. The line segments are discretized into $K$ points to form $\mathcal{C}_0(w_t)$. Given the corresponding gripper pose $T_t\in SE(3)$, CER is $\mathcal{C}(T_t,w)=T_t\mathcal{C}_0(w_t)$, and the canonicalized policy input is
\begin{equation}
    \tilde{P}_t
    =
    P_t^{\mathrm{scene}}
    \cup
    \mathcal{C}(T_t,w_t).
    \label{eq:cer}
\end{equation}

The same CER design is used for both human and robot demonstrations. For robot demonstrations, $T_t$ and $w_t$ are obtained directly from the gripper pose and width, respectively. For human demonstrations, following prior practices~\cite{lidea,phantom,masquerade}, we construct a gripper-aligned frame from hand landmarks: the thumb-index direction defines the opening axis, while their midpoint-to-palm direction defines a second axis. These landmarks determine $T_t$ and $w_t$, from which we instantiate the same CER.

While we instantiate CER with gripper geometry in this work, the formulation is not specific to grippers: $\mathcal{C}_0(s)$ can in principle be replaced by a canonical geometry for another end effector given a corresponding task-space pose mapping.

\begin{figure*}
    \centering
    \includegraphics[width=\linewidth]{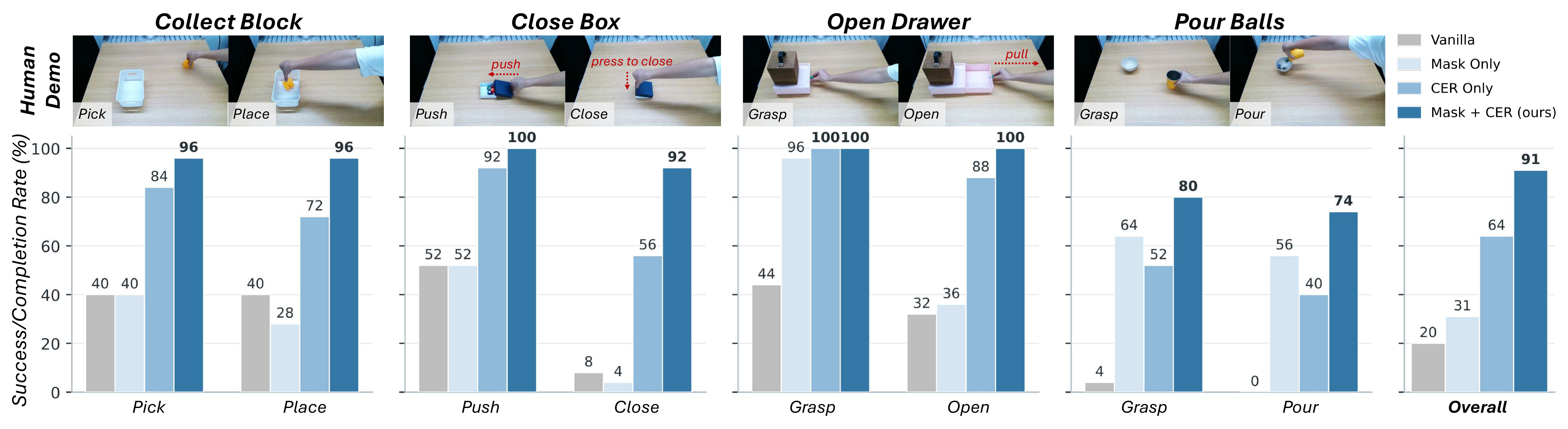}
\caption{
\textbf{Human-to-Robot Policy Transfer: Tasks and Results.}
\textit{\textbf{(Top)}} Illustration of the four real-world manipulation tasks and their task stages.
\textit{\textbf{(Bottom)}} Success/completion rates of robot policies trained without robot demonstrations. Embodiment canonicalization consistently improves human-to-robot policy transfer across diverse manipulation behaviors.
}
    \label{fig:human}\vspace{-0.45cm}
\end{figure*}

\subsection{Configuration-Decorrelation Augmentation}
\label{sec:method-aug}

A key property of CER is that its configuration can be directly edited in 3D. We exploit this property to diversify associations between task state and embodiment configuration beyond those observed in successful demonstrations. We apply augmentation only to \textit{recovery-eligible} states, where perturbing the gripper does not alter the underlying task state and the demonstrated behavior remains recoverable. Typical examples include free-space approach and pre-grasp states, while contact-constrained states are excluded when changing the gripper pose would require modifying the object or surrounding scene. For an eligible state at time $t$, we sample a bounded rigid-body perturbation $\Delta T_t$ and construct $\tilde{T}_t = \Delta T_t T_t$, where $T_t$ is the gripper pose in the demonstration. The corresponding augmented observation is
\begin{equation}
    P_t^{\mathrm{aug}}
    =
    P_t^{\mathrm{scene}}
    \cup
    \mathcal{C}(\tilde{T}_t, w_t).
    \label{eq:aug}
\end{equation}
Since the original embodiment has already been removed, the CER configuration can be modified without re-rendering the robot or changing the surrounding task scene.

Each perturbed configuration is paired with a corresponding recovery trajectory rather than the original demonstrated action. We interpolate a task-space trajectory from $\widetilde T_t$ back to the demonstrated pose $T_t$, using linear interpolation for translation and interpolation in $SO(3)$ for orientation. We retain only perturbations whose poses and recovery trajectories satisfy workspace and collision constraints. Once the gripper reaches $T_t$, the trajectory continues along the original demonstration, \textit{i.e.}, $\tilde{\tau} = [\tilde{T}_t \stackrel{\text{interp}}{\longrightarrow} T_t, T_{t+1}, \ldots]$.

The augmentation is performed on the fly during training. For each sample at a recovery-eligible state, we apply configuration-decorrelation augmentation with probability $p_{\mathrm{aug}}=0.5$; otherwise, the original sample is retained. When applied, a perturbation is randomly sampled to construct the augmented observation and recovery trajectory. Repeated resampling exposes the policy to diverse associations between task state and embodiment configuration.

\section{Experiments}\label{sec:experiments}

\begin{figure*}
    \centering
    \includegraphics[width=\linewidth]{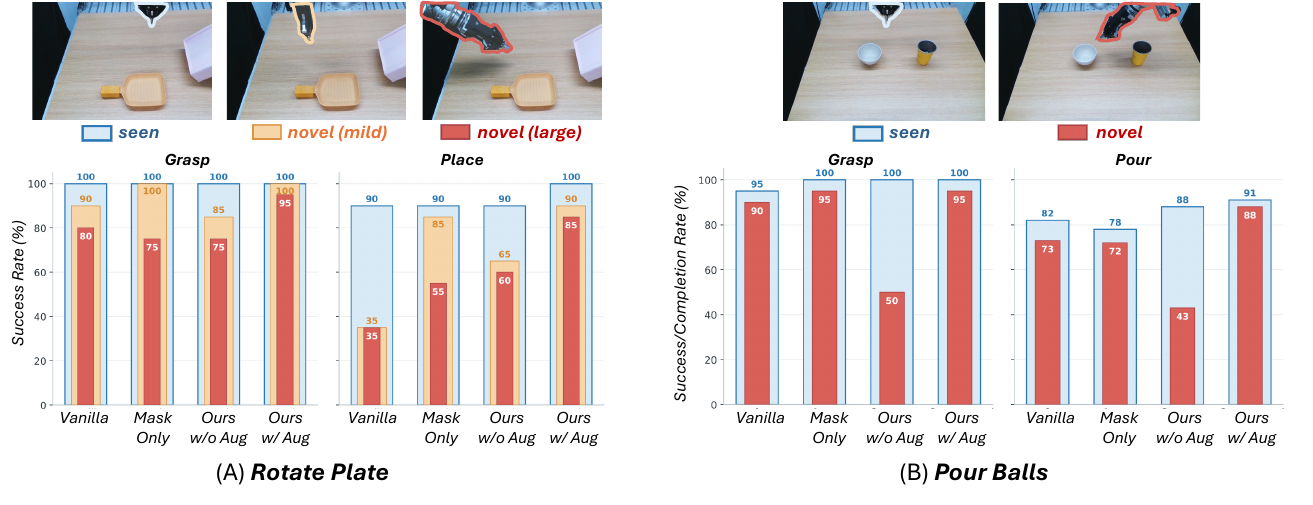}\vspace{-0.3cm}
    \caption{
        \textbf{Generalization to Novel Robot Configurations.}
        \textit{(A)} \textbf{\textit{Rotate Plate}} under seen, mild, and large configuration shifts.
        \textit{(B)} \textbf{\textit{Pour Balls}} under seen and novel configurations.
        While CER improves control under seen configurations, it can remain brittle when robot configuration becomes decoupled from task progress; configuration-decorrelation augmentation restores robustness without sacrificing seen performance.
    }
    \label{fig:generalization}\vspace{-0.45cm}
\end{figure*}

We conduct all experiments on a Flexiv Rizon 4 robot with a Dahuan AG-95 gripper. A fixed Intel RealSense D415 RGB-D camera provides global visual observations. Our experiments aim to address the following questions:
\textbf{(Q1)} How does embodiment canonicalization support human-to-robot policy transfer, and what roles do morphology suppression and control grounding play?
\textbf{(Q2)} How does embodiment canonicalization affect generalization to unfamiliar robot configurations, and can configuration-decorrelation augmentation improve robustness?
\textbf{(Q3)} What embodiment information should be suppressed and preserved for effective cross-embodiment learning?
\textbf{(Q4)} What deployment overhead does embodiment masking introduce?
\textbf{(Q5)} Does embodiment canonicalization redirect learned representation dependence from original morphology toward CER?

\subsection{Human-to-Robot Policy Transfer}
\label{sec:exp-human-videos}

\textbf{Tasks.}
As shown in Fig.~\ref{fig:human}(top), we design 4 tasks covering diverse manipulation behaviors: pick-and-place (\textbf{\textit{Collect Block}}), non-prehensile pushing (\textbf{\textit{Close Box}}), articulated object manipulation (\textbf{\textit{Open Drawer}}), and orientation-intensive manipulation (\textbf{\textit{Pour Balls}}). The acting hand remains visually prominent during manipulation, creating substantial morphology differences from the robot and making these tasks suitable for evaluating cross-embodiment visual transfer.

\textbf{Data.}
We collect 50 human demonstrations for each task. Policies are trained from scratch exclusively on these human demonstrations, without any robot demonstration data. AnyHand-trained WiLoR~\cite{wilor,anyhand} is used for hand detection and tracking, and POEMv2~\cite{poemv2} recovers frame-wise 3D hand landmarks. We use these landmarks to construct the gripper-aligned frame for CER instantiation (\S\ref{sec:method-cer}), and SAM2~\cite{sam2} to obtain embodiment masks. The retargeted actions are further processed with temporal filtering and orientation regularization to suppress noisy pose variations. As our focus is the visual embodiment gap rather than action retargeting, we use the same action-processing pipeline for all methods and do not study it further.

\textbf{Baselines.}
Most prior cross-embodiment canonicalization methods operate on 2D images and are coupled with different policy backbones~\cite{shadow,phantom,masquerade}. Directly comparing them with our 3D formulation would confound embodiment processing with observation modality and policy architecture. Instead, we use controlled variants of the same RISE backbone, including ``Vanilla'' without processing, ``Mask Only'' that applies embodiment masking, analogous to~\cite{shadow}, and ``CER Only'' that overlays CER without masking, to isolate the effects of our design. See \S\ref{sec:exp-ablation} for discussion of embodiment-translation methods~\cite{phantom,masquerade} and their ideal performance.

\textbf{Protocols.}
Policies run on a workstation equipped with an NVIDIA RTX 3090 GPU during inference. For evaluation, we randomly sample 25 initial object configurations within the workspace and use the same set for all policies. Each policy is evaluated once on each configuration, and we report both phase-wise and overall success rates. For \textbf{\textit{Pour Balls}}, we additionally report the completion rate of the ``pour'' stage as the fraction of balls successfully poured into the bowl.

\textbf{Embodiment canonicalization consistently improves human-to-robot policy transfer by jointly suppressing morphology-specific cues and preserving control grounding (Q1).}
As shown in Fig.~\ref{fig:human}, embodiment canonicalization achieves the best overall performance on all four tasks, increasing the average success rate from 20\% to 91\%, compared with 31\% for ``Mask Only'' and 64\% for ``CER Only''. Masking alone can reduce morphology dependence but lacks explicit end-effector geometry for control grounding, while CER provides such geometry but remains less effective when the original human morphology is still visible. Their consistent gap to embodiment canonicalization shows that effective human-to-robot transfer benefits from combining morphology suppression with control-relevant geometry.

\textbf{Preserving control-relevant geometry becomes particularly important for precise or sustained end-effector control (Q1).}
On \textbf{\textit{Open Drawer}}, ``Mask Only'' substantially improves grasping from 44\% to 96\%, yet barely improves opening from 32\% to 36\%, whereas CER substantially improves the constrained opening motion. \textbf{\textit{Pour Balls}} further stresses precise control grounding because the cup width is close to the gripper's maximum opening, leaving little tolerance for grasp-position or orientation errors. ``Mask Only'' and ``CER Only'' often produce visibly misaligned grasps, while embodiment canonicalization achieves the highest grasp and pouring success. Together, these results show that robust cross-embodiment transfer requires not simply removing embodiment information, but preserving precise, transferable control-relevant geometry.
\subsection{Generalization to Novel Robot Configurations}
\label{sec:exp-novel-config}

\textbf{Tasks and Data.}
We evaluate generalization to novel robot configurations on \textbf{\textit{Rotate Plate}} (Fig.~\ref{fig:diagnostic} in \S\ref{sec:ved_diagnostic}) and \textbf{\textit{Pour Balls}} (Fig.~\ref{fig:human} in \S\ref{sec:exp-human-videos}). Both tasks involve substantial end-effector rotation and consequently large variations in visible robot configuration, making them suitable for studying configuration dependence. We collect 50 teleoperated demonstrations for each task using haptic devices~\cite{rh20t}. The \textbf{\textit{Rotate Plate}} demonstrations are also used for the VED diagnostic in \S\ref{sec:ved_diagnostic}.

\textbf{Protocols.}
We follow the evaluation protocol in \S\ref{sec:exp-human-videos} and sample 20 initial object configurations for each task. To evaluate robot-configuration generalization, we sample robot configurations from the demonstration distribution and deliberately pair them with the \textit{initial task state}, creating configurations that are decoupled from the task progress observed during training. For \textbf{\textit{Rotate Plate}}, we further divide these configurations into \textit{mild} and \textit{large} shifts according to their deviation from the demonstrated initial configurations. For \textbf{\textit{Pour Balls}}, we evaluate a single held-out configuration shift and report the completion rate of the \textit{pour} stage. All policies are evaluated on the same object initializations and robot configurations for fair comparison.

\textbf{Embodiment canonicalization alone can remain brittle to unfamiliar robot configurations, while configuration-decorrelation augmentation consistently improves robustness (Q2).}
As shown in Fig.~\ref{fig:generalization}, masking alone can already improve robustness under moderate shifts, but its benefit becomes limited as the configuration mismatch increases. More importantly, CER exhibits a clear tension: it improves control under seen configurations, yet can become substantially more brittle when robot configuration becomes decoupled from task progress, particularly on \textbf{\textit{Pour Balls}}. This suggests that \textit{canonicalizing morphology does not by itself resolve configuration dependence}, since CER can still become tightly coupled with demonstrated task progress. Configuration-decorrelation augmentation addresses this failure mode by exposing the policy to off-trajectory CER configurations and corresponding recovery behaviors, consistently recovering strong performance across both tasks and substantially improving pouring completion under the novel configuration. These results show that robust configuration generalization requires not removing configuration information, but diversifying how configuration is associated with behavior.

\subsection{Ablations and Analysis}\label{sec:exp-ablation}

\begin{figure*}
\centering
\begin{minipage}{0.63\textwidth}
    \centering
    \includegraphics[width=0.95\linewidth]{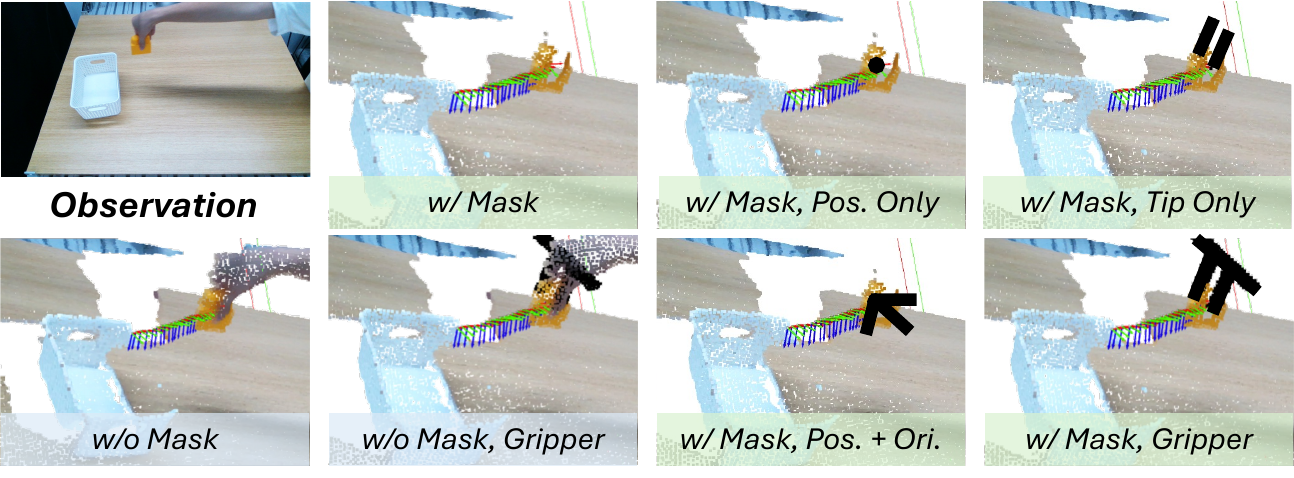}
\end{minipage}\hfill
\begin{minipage}{0.37\textwidth}
    \centering\footnotesize
    \begin{tabular}{clrrrr}
        \toprule
        \multirow{2}{*}{\textbf{Mask}} & \multirow{2}{*}{\textbf{CER Design}} & \multicolumn{2}{c}{\textbf{Success Rate}} \\ \cmidrule(lr){3-4}
        & & Pick & Place  \\
        \midrule
        - & - & 40\% & 40\% \\
        -  & Gripper & 84\% & 72\% \\ \midrule
        \checkmark & - & 40\% & 28\% \\
        \checkmark & Pos. Only & 36\% & 24\% \\ 
        \checkmark & Pos. + Ori. & 64\% & 60\% \\
        \checkmark & Gripper Tip Only & 68\% & 68\% \\
        \rowcolor[HTML]{F2F2F2} \checkmark & Gripper (\textit{ours}) & \textbf{96}\% & \textbf{96}\% \\ \midrule
        \checkmark & Robot \textit{(oracle, via replay)} & \textbf{96}\% & \textbf{96\%} \\
        \bottomrule
    \end{tabular}
\end{minipage}\vspace{-0.1cm}
\captionof{table}{
    \textbf{Ablation of Mask and CER Design on the \textit{Collect Block} Task.}
    \textit{(Left)} Illustration of different CER designs and the processed point cloud for policy input. The coordinate frames are the ground-truth actions at this observation. \textit{(Right)} Ablation results show that masking alone or adding CER alone is insufficient, while progressively richer end-effector geometry improves cross-embodiment policy learning. Our full CER design matches the \textit{oracle} robot representation obtained via replaying, suggesting that the compact canonical geometry preserves the control-relevant information of the complete robot morphology.}\label{tab:ablation-cer}
    \vspace{-0.4cm}
\end{figure*}

\textbf{Both suppressing the original embodiment and preserving sufficiently expressive end-effector geometry are important for effective cross-embodiment transfer (Q3).}
As shown in Tab.~\ref{tab:ablation-cer}, CER Only already provides a substantial improvement over the vanilla policy, while embodiment canonicalization further achieves 96\% success on both task stages. Among the masked variants with different CER designs, richer geometric representations consistently outperform coarse position- or pose-based representations, with the full canonical gripper performing best. These results suggest that effective embodiment canonicalization should suppress embodiment-specific morphology while retaining sufficiently expressive end-effector geometry for control grounding.

\textbf{Our CER design can match an oracle representation with the full robot morphology (Q3).}
To estimate the gap between our CER design and an idealized robot-side representation, we \textit{replay} the retargeted trajectories on the robot, capture the corresponding full robot geometry, and use it in place of our CER design to construct an \textit{oracle} training set. Despite containing the complete robot morphology, this oracle achieves the same success rate as our CER design (Tab.~\ref{tab:ablation-cer}). Notably, such full-robot reconstruction approximates the ideal target of prior embodiment-translation methods~\cite{phantom,masquerade}. This suggests that our compact canonical gripper can retain the control-relevant visual information provided by the full embodiment without complete reconstructions.

\textbf{Deployment-time masking adds modest overhead and is not strictly required to retain most of the benefit (Q4).}
As shown in Tab.~\ref{tab:ablation-masking}, when the robot URDF and camera calibration are available, the mask can instead be rendered directly, maintaining 96\% success with only 9\,ms additional latency over vanilla inference, substantially lower than SAM2~\cite{sam2}. Interestingly, removing deployment-time masking causes only a small drop from 96\% to 92\%, while restoring inference latency to nearly the vanilla level. This motivates us to further examine how training-time embodiment masking affects the learned representation.

\begin{table}[t]
    \centering    \vspace{-0.1cm}
    \setlength\tabcolsep{2.5pt}
    \begin{tabular}{lllrr}
        \toprule
        \textbf{Training} & \textbf{Deployment} & \textbf{Mask Source} & \textbf{Success Rate} & \textbf{Latency} \\ \midrule
        Mask + CER & Mask + CER & SAM2~\cite{sam2} & 96\% & 237ms\\
        Mask + CER & Mask + CER & URDF Rendering & 96\% & 190ms \\
        Mask + CER & CER Only & - & 92\% & 182ms\\
        CER Only & CER Only & - & 72\% & 182ms\\
        Vanilla & Vanilla & - & 40\% & 181ms\\
        \bottomrule
    \end{tabular}
    \caption{\textbf{Ablation of Masking at Training and Deployment.}
Training with Mask+CER is important for learning embodiment-robust policies, while removing masking at deployment incurs only a modest performance drop on \textbf{\textit{Collect Block}}. If the robot URDF is available, rendering the mask reduces inference latency.}
    \label{tab:ablation-masking}\vspace{-0.6cm}
\end{table}

\begin{figure}[t]
    \centering
    \vspace{-0.2cm}
    \includegraphics[width=\linewidth]{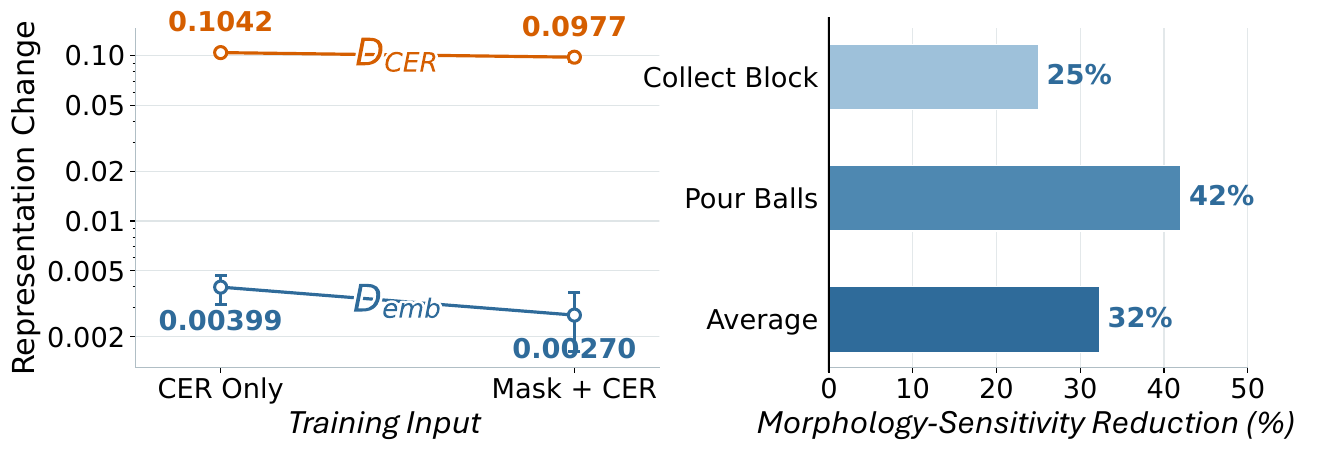}\vspace{-0.2cm}
    \caption{
\textbf{Representation Sensitivity Analysis.}
\textit{(Left)} Training-time masking reduces sensitivity to the original morphology $D_\text{emb}$.
\textit{(Right)} The attention to CER $D_\text{CER}$ remains substantial, indicating preserved dependence on control-relevant embodiment geometry.
}
    \label{fig:repr}\vspace{-0.5cm}
\end{figure}

\textbf{Training-time embodiment masking reduces dependence on the original morphology while preserving CER dependence (Q5).}
Using held-out robot observations from \textbf{\textit{Collect Block}} and \textbf{\textit{Pour Balls}}, we perform controlled interventions by independently removing the original embodiment or CER while keeping the task scene fixed. We extract the final-layer representation in RISE immediately before the action decoder and quantify representation changes using cosine distance. As shown in Fig.~\ref{fig:repr}, adding training-time masking reduces sensitivity to the original embodiment, while the focus on CER remains substantial. These results suggest that \textit{masking during training suppresses learned dependence on embodiment-specific morphology without eliminating the control-relevant embodiment information provided by CER}.

\section{Conclusion}\label{sec:conclusion}

In this paper, we revisit how visuomotor policies use visual embodiment information. We show that visual embodiment dependence is not simply something to remove: embodiment geometry may provide useful control grounding, while morphology-specific cues and configuration-progress correlations can become brittle shortcuts. Our results show that embodiment canonicalization and configuration-decorrelation augmentation can structure this dependence to improve human-to-robot transfer and robustness to unfamiliar robot configurations. Overall, the key question is not whether policies should depend on embodiment information, but \textit{which embodiment information they should depend on}.

\textbf{Limitations and Future Work.}
This work assumes that the end effector captures the primary control-relevant embodiment information. Current performance also depends on reliable masks. Future work could extend CER to dexterous hands and other embodiments, explore learning from more diverse human demonstrations and egocentric human videos, and generalize embodiment canonicalization beyond 3D point clouds to 2D and other observation modalities.

% not for ICRA, but can be included in arXiv version
\section*{Acknowledgement}

We would like to thank Peishen Yan from Shanghai Jiao Tong University for his suggestions on the paper writing.
 
\printbibliography
 
% not for ICRA, but can be included in arXiv version
\clearpage
\appendices

\section{Implementation Details}
\label{app:implementation}

\subsection{Policy Training}

We use RISE~\cite{rise} as the visuomotor backbone for all experiments. Unless otherwise specified, all policy variants use the same network architecture, training objective, optimization schedule, and data preprocessing pipeline, such that the only differences are the embodiment processing and augmentation strategies described in the main paper.

For point-cloud preprocessing, RGB-D observations are converted into colored 3D point clouds and voxelized with a voxel size of 5\,mm. We crop the point cloud to a fixed workspace of $[-0.5\text{m}, 0.5\text{m}]\times [-0.6\text{m}, 0.4\text{m}] \times [0.3\text{m}, 1.3\text{m}]$ in the camera frame. The policy is trained for 1000 epochs (approximately 30000 steps) with a batch size of 240, using AdamW optimizer with a learning rate of $3\times 10^{-4}$. We use single-step point-cloud observations only, without proprioception, to predict the 20-step action chunk in the camera frame. These hyperparameters follow~\cite{rise}.

\subsection{Embodiment Masks}

For human demonstrations, we use SAM2~\cite{sam2} to obtain embodiment masks. For robot observations, the embodiment mask is obtained using SAM2. We dilate the mask using a kernel size of 10 to account for segmentation errors before removing the corresponding depth points. At deployment, when the robot URDF and camera calibration are available, the robot mask can be rendered directly from the current robot state. As shown in Tab.~\ref{tab:ablation-masking}, this reduces the additional latency of deployment-time masking to 9\,ms over vanilla inference while maintaining the same success rate as SAM2.

\subsection{CER Construction}
\label{app:cer}

We instantiate CER as a lightweight pseudo-gripper represented directly as a 3D point cloud. The canonical geometry consists of three modules: two finger modules and one proximal gripper-body module. Each finger is represented by a rectangular volume of size $2\,\mathrm{cm}\times1\,\mathrm{cm}\times 6\,\mathrm{cm}$, while the proximal module has a thickness of $1\,\mathrm{cm}$ and spans the full gripper range. The geometry is initially sampled densely and voxel-downsampled at 5\,mm resolution, consistent with the point-cloud preprocessing used by the policy.

CER additionally preserves the gripper opening width. Given the current opening width $w_t$, the two finger modules are translated symmetrically to $-w_t/2$ and $+w_t/2$ along the opening axis. We then write the canonical template as $\mathcal{C}_0(w_t)$ and instantiate CER at end-effector pose
$T_t\in SE(3)$ as
\[
    \mathcal{C}(T_t,w_t)
    =
    T_t\mathcal{C}_0(w_t).
\]
The resulting points are assigned a uniform black RGB feature and appended
to the task-scene point cloud.

\subsection{Human Demonstration Processing}

We collect 50 human demonstrations for each task: \textbf{\textit{Collect Block}}, \textbf{\textit{Close Box}}, \textbf{\textit{Open Drawer}}, and \textbf{\textit{Pour Balls}}. No robot demonstrations are used for training the policies evaluated in human-to-robot transfer. The recovered human hand motion is retargeted to the robot end-effector space. We apply temporal filtering and orientation regularization to suppress noisy frame-to-frame variations. The same retargeting and action-processing pipeline is used for all compared methods.

\section{Configuration-Decorrelation Augmentation}
\label{app:augmentation}

\subsection{Recovery-Eligible States}

Configuration-decorrelation augmentation is applied only at \textit{recovery-eligible} states, where perturbing the end-effector does not change the underlying task state and the demonstrated behavior remains recoverable. Typical examples include free moving and pre-grasp states. We exclude contact-constrained states when changing the gripper pose would require modifying the object or surrounding task scene.

\subsection{Perturbation and Recovery Generation}

For an eligible sample at time $t$, we sample a bounded rigid-body perturbation $\Delta T_t$ and construct $\widetilde{T}_t=\Delta T_tT_t$. Translation perturbations are sampled within the workspace, and rotation perturbations are sampled within $60^\circ$.

The augmented observation is constructed by replacing the original CER with $\mathcal{C}(\widetilde{T}_t)$ while keeping the task scene unchanged. We then generate a recovery trajectory from $\widetilde{T}_t$ back to the pose $T_t$, constrainted by the linear and angular velocities $v_\text{max} = 0.02\,\text{m/step}, \omega_\text{max} = 0.15\,\text{rad/step}$. Translation is linearly interpolated, while orientation is interpolated in $SO(3)$ using spherical linear interpolation (SLERP). Perturbations whose poses or recovery trajectories violate workspace or collision constraints are rejected.

Once the gripper reaches $T_t$, the recovery trajectory reconnects to the original demonstration:
\[
\widetilde{\tau}
=
[\widetilde{T}_t
\stackrel{\mathrm{interp}}{\longrightarrow}
T_t,T_{t+1},\ldots].
\]

The augmentation is sampled on the fly with probability
$p_{\mathrm{aug}}=0.5$ for each recovery-eligible sample.

\section{Additional Details of the VED Diagnostic}
\label{app:ved}

\subsection{Cue-Conflict Construction}

For two task stages $i$ and $j$, we denote the observations as $o_i=(E_i,S_i)$ and $o_j=(E_j,S_j)$, where $E$ is the visible embodiment configuration and $S$ denotes the remaining visual task state. The cue-conflict observation is $o_{i\rightarrow j}=(E_j,S_i)$. In simulation, we restore task state $S_i$, set the robot to configuration $E_j$, and re-render the observation. In the real world, we physically reproduce task state $S_i$, move the robot to configuration $E_j$, and capture a new RGB-D observation. This avoids image-level copy-paste artifacts and produces physically consistent cue conflicts. For each selected stage pair, we construct $N=20$ cue-conflict observations. The same observations are evaluated by all policies. We only use stage pairs whose task-implied and embodiment-implied absolute
actions differs substantially

\subsection{Action Distance}

For absolute end-effector actions
$a=(p,R)$, we define the action distance as
\[
d(a_1,a_2)
=
\|p_1-p_2\|_2
+
\lambda_R
d_{\mathrm{SO}(3)}(R_1,R_2),
\]
where $d_{\mathrm{SO}(3)}$ is the geodesic rotation distance in radians. We set $\lambda_R=0.06\,\mathrm{m/rad}$, corresponding to the approximate gripper length, to convert rotational error to an equivalent spatial displacement at the end effector. The gripper command is excluded from the diagnostic distance.

\section{Full-Robot Oracle Construction}
\label{app:oracle}

To estimate the gap between our compact CER design and an idealized robot-side representation, we construct a full-robot oracle on \textbf{\textit{Collect Block}}. Starting from the retargeted human trajectories, we replay the corresponding end-effector motions on the robot and capture the resulting full robot geometry. The captured robot geometry replaces CER during training, while the remaining task scene and action-processing pipeline are kept unchanged. This produces an oracle representation in which the human embodiment is replaced by the complete deployment-time robot morphology.

This construction approximates the ideal target representation of embodiment-translation approaches that aim to transform human observations into robot-like observations~\cite{phantom,masquerade,airexo2_rise2}. The oracle achieves 96\% success on both \textit{Pick} and \textit{Place} stages, matching our compact CER design despite retaining the complete robot morphology.

\section{Representation Sensitivity Analysis}
\label{app:representation}

We analyze 42 held-out robot observations: 17 frames from three \textbf{\textit{Collect Block}} rollouts and 25 frames from three \textbf{\textit{Pour Balls}} rollouts. Both ``CER Only'' and ``Mask+CER'' policies are evaluated on the same frames, without retraining or result-based frame selection.

For each frame, we construct four variants: \textbf{Full}, with the original robot geometry; \textbf{Mask}, with the robot geometry removed; \textbf{Full+CER}; and \textbf{Mask+CER}. All variants share the same RGB-D frame, task scene, robot state, and CER pose, and are constructed before point-cloud preprocessing.

We extract the final Transformer readout in RISE immediately before the diffusion action decoder and measure representation changes using cosine distance. Original-morphology and CER sensitivities are defined as
\begin{equation}
\left\{
    \begin{aligned}
        D_{\mathrm{emb}} &=1-\cos\left(z_{\mathrm{Full+CER}}, z_{\mathrm{Mask+CER}}\right),\\
        D_{\mathrm{CER}} &=1-\cos\left(z_{\mathrm{Mask+CER}},z_{\mathrm{Mask}}\right).
    \end{aligned}
\right.
\end{equation}

Confidence intervals (CIs) are estimated with $10^4$ bootstrap replicates over complete rollouts. As shown in Tab.~\ref{tab:repr-full}, training-time masking reduces morphology sensitivity by approximately 32\% overall, with a paired difference of $-0.00129$ and a 95\% CI of $[-0.00178,-0.00082]$. The same trend appears on both tasks, while sensitivity to CER remains substantial. These results indicate that training-time masking suppresses dependence on embodiment-specific morphology while preserving dependence on the control-relevant geometry provided by CER.

\begin{table}[h]
    \centering
    \setlength\tabcolsep{4pt}
    \begin{tabular}{llcc}
        \toprule
        \textbf{Task} & \textbf{Training} &
        $D_{\mathrm{emb}}$ &
        $D_{\mathrm{CER}}$ \\
        \midrule
        Collect Block & CER Only & 0.00558 & 0.09085 \\
                      & Mask+CER & 0.00417 & 0.09813 \\
        \midrule
        Pour Balls    & CER Only & 0.00290 & 0.11324 \\
                      & Mask+CER & 0.00170 & 0.09746 \\
        \midrule
        Overall       & CER Only & 0.00399 & 0.10417 \\
                      & Mask+CER & 0.00270 & 0.09773 \\
        \bottomrule
    \end{tabular}
    \caption{\textbf{Representation Sensitivity.}
    Training-time masking reduces sensitivity to original morphology while preserving substantial sensitivity to CER. Values are cosine distances averaged over held-out robot observations.}
    \label{tab:repr-full}
\end{table}

\end{document}